\documentclass[sigconf]{acmart}

\AtBeginDocument{%
  }

\setcopyright{acmlicensed}
\copyrightyear{2024}
\acmYear{2024}
\acmDOI{XXXXXXX.XXXXXXX}

\usepackage{tcolorbox}
\usepackage{multicol}
\usepackage{multirow}
\usepackage{tabularx}
\usepackage{adjustbox}
\usepackage{arydshln}
\usepackage{xcolor}
\usepackage{bbding}
\usepackage{enumitem}
\usepackage{balance}
\usepackage{pifont}
\usepackage[normalem]{ulem}

\begin{document}
\setlength{\parindent}{0em}

\title{\textit{Let Silence Speak:} Enhancing Fake News Detection with Generated Comments from Large Language Models}

\author{Qiong Nan\textsuperscript{1,2}, Qiang Sheng\textsuperscript{1}, Juan Cao\textsuperscript{1,2}, Beizhe Hu\textsuperscript{1,2}, Danding Wang\textsuperscript{1}, Jintao Li\textsuperscript{1}}
\affiliation{%
\institution{\textsuperscript{1}Institute of Computing Technology, Chinese Academy of Sciences \\
\textsuperscript{2}University of Chinese Academy of Sciences
}
\streetaddress{}
\city{}
\state{}
\country{}
\postcode{}
}
\email{{nanqiong19z, shengqiang18z, caojuan, hubeizhe21s, wangdanding, jtli}@ict.ac.cn}

\renewcommand{\shortauthors}{Nan et al.}
\begin{abstract}
Fake news detection plays a crucial role in protecting social media users and maintaining a healthy news ecosystem.
Among existing works, comment-based fake news detection methods are empirically shown as promising because comments could reflect users' opinions, stances, and emotions and deepen models' understanding of fake news.
Unfortunately, due to exposure bias and users' different willingness to comment, it is not easy to obtain diverse comments in reality, especially for early detection scenarios.
Without obtaining the comments from the ``silent'' users, the perceived opinions may be incomplete, subsequently affecting news veracity judgment.
In this paper, we explore the possibility of finding an alternative source of comments to guarantee the availability of diverse comments, especially those from silent users.
Specifically, we propose to adopt large language models (LLMs) as a user simulator and comment generator, and design \textbf{GenFEND}, a generated feedback-enhanced detection framework, which generates comments by prompting LLMs with diverse user profiles and aggregating generated comments from multiple subpopulation groups.
Experiments demonstrate the effectiveness of GenFEND and further analysis shows that the generated comments cover more diverse users and could even be more effective than actual comments.
\end{abstract}



\maketitle

\section{Introduction}
\begin{figure}[t]
    \centering
    \includegraphics[width=1.0\linewidth]{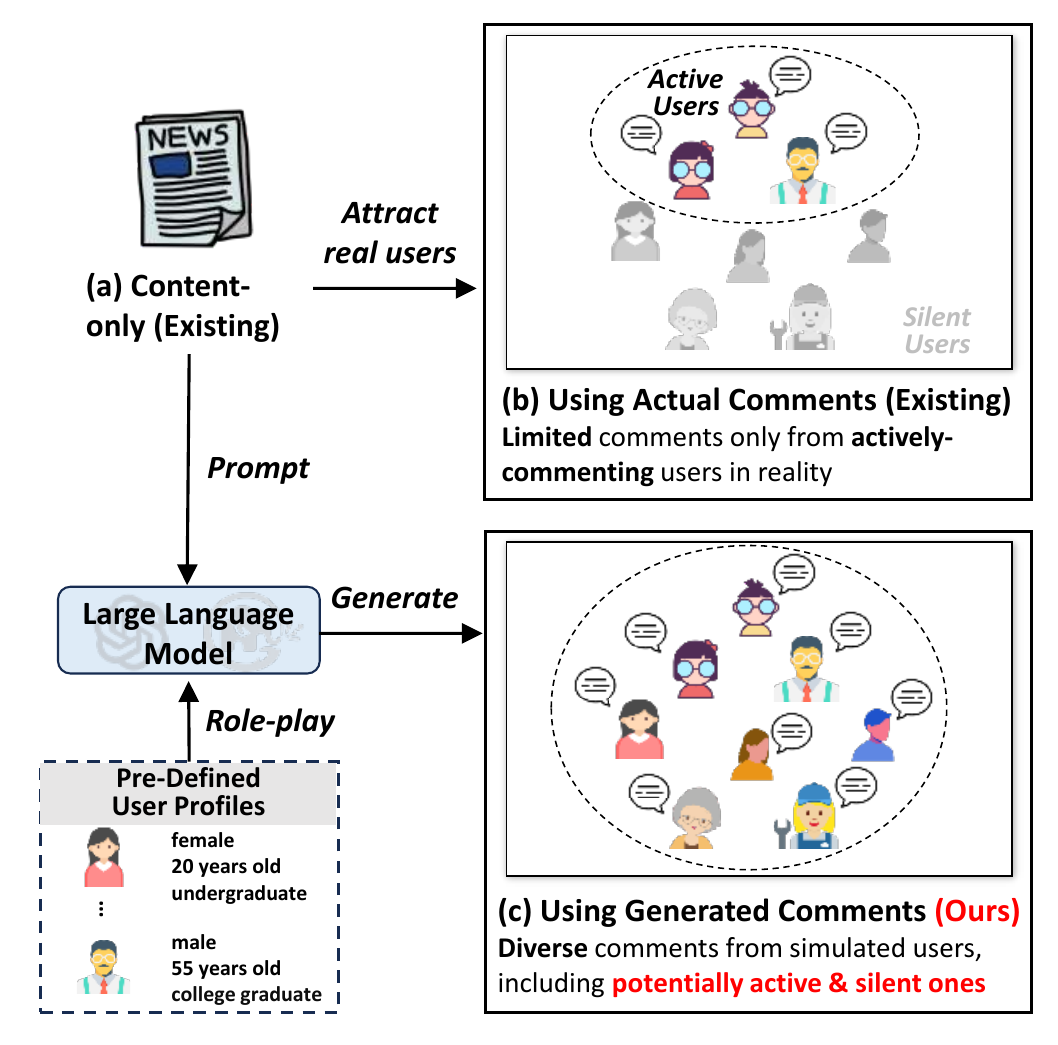}
    \caption{\label{fig:fig1} Existing fake news detection methods rely on (a) the news content itself and (b) limited comments from actively commenting users only. Unlike (a) and (b), our GenFEND uses (c) diverse comments generated by large language models from both potentially active and silent simulated users.}
    \vspace{-0.3cm}
\end{figure}
The rapid spread of fake news on social media platforms can lead to substantial losses within a short period~\cite{Vosoughi_Science2018, Tasnim_ImpactCOVID19}. 
For example, a fake news post reporting an explosion in the White House triggered panic and caused the Dow Jones index to fall 100 points in just two minutes~\cite{obama}.
More seriously, the malicious use of large language models (LLMs) facilitates fake news creation and may bring larger risks in the near future~\cite{global-risk,Chen_LLM_Mis,jiang2024disinformation,barman2024dark}.
Although human countermeasures like establishing reporting mechanisms~\cite{weibo-convention} and conducting fact-checking~\cite{walter2020fact} have been adopted, their inevitable lagged effect makes it hard to achieve the ultimate goal of moderating fake news as early as possible. Therefore, recent studies have focused on automatic fake news detection.

Existing fake news detection methods generally rely on analyzing news content or introducing external resources as references~\cite{Shu_FakeNewsSurvey}. Among the resources, comments from social media users play a valuable role and have been shown promising helpfulness~\cite{Nan_CASFEND}.
The unique advantage of comment-based methods can be attributed to the support of \textit{crowd intelligence}, which reflects various users' understanding of news, such as opinions~\cite{Yang_UFD}, stances~\cite{Ma_RumorStance}, and emotions~\cite{Zhang_DualEmo}. By perceiving and aggregating the patterns behind crowd signals, detectors would differentiate real and fake news more easily.
Unfortunately, maintaining the quantity and quality of user comments is hard in real-world scenarios due to the following reasons: 
(1) At the early stage of news dissemination, it is unlikely to attract a wide audience to comment. 
(2) Even if after a period of dissemination, the available comments only reflect opinions from partial, active user groups that are willing to make comments, due to the intrinsic commenting unwillingness of specific users. 
For example, comments from users with higher degrees and professional knowledge may help detect misleading scientific news, but such users may rarely view and comment on suspicious posts.
(3) Moreover, the distribution of available comments is often unstable due to the exposure bias influenced by factors like issued time, recommender system preferences, etc., making it harder for detectors to mine clear, stable patterns. For example, a memorized pattern may be ineffective due to formally active users' inactivation and key comments missing.
Therefore, with limited comments, existing comment-based detectors can only have a limited observation and biased understanding of crowd feedback, which ultimately risks the detection performance.
\textbf{It is valuable to find a surrogate of comments from real users that comprehensively reflects crowd intelligence and facilitates a deep news understanding in fake news detection.}

\textbf{In this paper, we leverage comments generated from large language models (LLMs) as an alternative.}
LLMs like ChatGPT possess impressive capabilities in natural language understanding and generation~\cite{wei2022emergent,Wang_DEEM,vats2024exploring,guo2024integrating}. 
Moreover, LLMs can simulate user behaviors following specific instructions in various applications, such as dialogue~\cite{Li_LLMFairness,Xie_SimHumanTrust,sekulić_SimulatorDialogue} and recommender systems~\cite{huang2024large}. 
Along this line, we prompt LLMs to generate diverse comments by role-playing different users, as depicted in Figure~\ref{fig:fig1}. 
To use generated comments for detection enhancement, we address two key challenges: (1) How to generate diverse comments using LLMs? (2) How to utilize the comments effectively?

To tackle these challenges, we propose a Generated Feedback Enhanced Detection (\textbf{GenFEND}) framework, which enhances fake news detection performance regardless of the availability of actual comments by real users.
we pre-define different user profiles by combining the attributes of gender, age, and education, according to which we prompt LLMs to generate diverse comments by role-playing these users to get comprehensive user feedback.
After extracting the semantic features of all generated comments, we split them into multiple subpopulation groups under each demographic view.
We further perform an average operation in each subpopulation group to get overall feedback and calculate the subpopulation-level divergences to represent differences. 
The final representation of generated comments is obtained via intra-view and inter-view aggregations.
Experiments demonstrate the effectiveness of GenFEND to enhance fake news detection performance.
Our main contributions are as follows:
\begin{itemize}[nosep,leftmargin=1em,labelwidth=*,align=left]
    \item \textbf{Idea:} We propose to induce LLMs to role-play social media users to generate diverse comments as a substitute for actual-posted ones for fake news detection.
    \item \textbf{Framework:} We design GenFEND, a generated feedback enhanced fake news detection framework, which generates diverse user comments, analyzes them from a multi-subpopulation perspective, and aggregates the derived features from both intra- and inter-demographic views.
    \item \textbf{Effects:} Experiments demonstrate the effectiveness of GenFEND to enhance fake news detection performance and show the unique value of LLM-generated comments. The code is available at \url{https://github.com/ICTMCG/GenFEND}.
\end{itemize}

\section{Related Work}
\textbf{Fake News Detection.} 
According to the sources of information used, methods for fake news detection can be clustered into two groups: content-based, and external resource-based methods~\cite{HU_FakeNewsSurvey}. 
Content-based methods often extract style features~\cite{Przybyla_AAAI2020_Capturing, Zhu_M3FEND}, semantic features~\cite{Yu_CNN_Misinfo, Zhang_Heterogeneous,hu-etal-2023-learn}, emotion features~\cite{ajao,Zhang_DualEmo} from textual content and visual features from appended images or videos in some multi-modal approaches~\cite{Singhal_SpotFake, Chen_Causal_Multimodal, Qi_Entity, Qi_FakeSV}. 
Recently, many researchers have investigated the (multi-modal) LLMs' capability as auxiliary tools for fake news detection, because of their remarkable performance in real-world knowledge understanding and reasoning~\cite{Hu_ARG, teller, Xu_LLMBelief, Liu_LLMDetect, Xuan_LEMMA, Wang_MMIDR, Wang_LLMDefense}.
We focus on the textual information in this paper. 
However, the performance of content-based methods is often limited by their vulnerability to fake news that intentionally mimics the style of real news~\cite{Gelfert_FakeNewsDefinition,wu2023fake}.
An alternative solution is to leverage auxiliary information from external resources, such as users~\cite{Shu_WeakSupervision, Liu_FNED, Castillo_InfoCred}, comments~\cite{Zhang_DualEmo, Ma_RumorStance, Shu_WeakSupervision, Jin_Viewpoints},  propagation structures~\cite{Liu_FNED}, news environments~\cite{Sheng_NEP}, and extra knowledge~\cite{Kou_HC-COVID, Yang_CoarsetoFine} because they introduce the crowd's help and provide related useful knowledge.
User comments are widely used because they are more informative than user attributes and easier to obtain compared to propagation structures and extra knowledge.
Different individuals' viewpoints from user comments can help identify fake news.
However, the performance of comment-based methods is largely influenced by the quantity and quality of actual comments, especially in early detection scenarios. 
As an alternative, prompting LLMs to generate reactions is initially applied in recent works~\cite{Liu_Skepticism, Wan_DELL}.
Following this line, we propose using LLMs to generate rich and diverse comments as a substitute for actual ones and enable a more comprehensive fake news understanding.

\begin{figure*}[t]
    \centering
    \includegraphics[width = \linewidth]{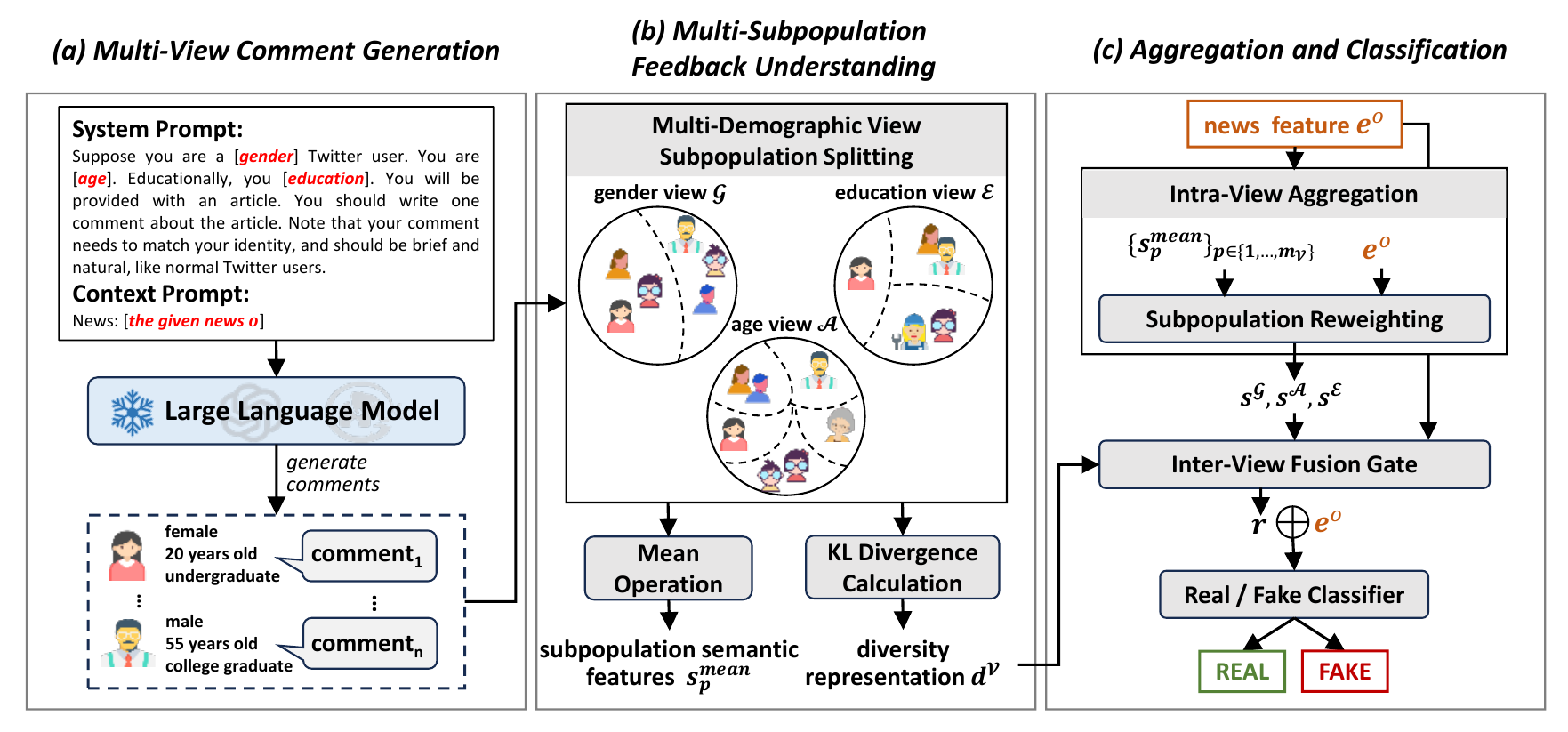}
    \caption{Overview of Generated Feedback Enhanced Detection (GenFEND) framework. (a) \textbf{Multi-View Comment Generation}: Pre-define different user profiles with three demographic characteristics (gender, age, and education); Then, prompt the LLM to generate comments by role-playing these users. (b) \textbf{Multi-Subpopulation Feedback Understanding}: Split generated comments into different subpopulation groups for each view; Extract the semantic feature $\boldsymbol{s}^{mean}_p$ for each subpopulation group $p$ and diversity representation $\boldsymbol{d}^\mathcal{V}$ for each view $\mathcal{V}$; (c) \textbf{Aggregation and Classification}: Perform intra-view aggregation by operating dot-product between semantic features $\{\boldsymbol{s}^{mean}_p\}_{p \in \left\{1, ..., m_\mathcal{V}\right\}}$ in each view $\mathcal{V}$ and news feature $\boldsymbol{e}^o$; Perform inter-view aggregation to get final feature $\boldsymbol{r}$ of generated comments with a fusion gate guided by news feature $\boldsymbol{e}^o$ and diversity representation $\boldsymbol{d} = \oplus_{\mathcal{V} \in \left\{\mathcal{G}, \mathcal{A}, \mathcal{E}\right\}}\boldsymbol{d}^\mathcal{V}$ as input; Concatenate $\boldsymbol{r}$ and $\boldsymbol{e}^o$ (and $\boldsymbol{e}^c_{actual}$ if available) for classification.}
    \label{fig:model}
\end{figure*}

\textbf{Comment Generation for News-Related Applications.} 
Comment generation is a sub-task to automatically produce human-like texts that express opinions on a given object. 
In news-related applications, comment-generation techniques can help synthesize pseudo comments to replace or supplement human-written comments and support social bot services for a more interactive discussion.
\citet{Yang_ReadAttendComment} modeled a ``read-attend-comment'' procedure with an encoder-decoder framework to generate comments more attentive to the key points in news articles.
\citet{Wang_Diversified} unified reader-aware topic modeling and saliency detection to improve the generation quality.
\citet{Zou_Controllable} adopted an attribute-level contrastive learning method to better control the mentioned elements.
Specified to fake news detection, \citet{Le_MALCOM} and \citet{Liang_COMCP} generated comments as a malicious attack on detection models.
\citet{Yanagi_GenCmts} generated comments with news articles and actual comments provided to replace actual ones.
Again, recent studies explored instructing LLMs to generate highly readable and human-like responses~\cite{Yu_PopALM,Sun_SocialSense,Wang_DEEM,Li_LLMFairness,sekulić_SimulatorDialogue}.
A contemporary work, DELL~\cite{Wan_DELL}, leverages LLMs to generate comments for social graph simulation and assists in graph-based fake news detection.
Our proposed GenFEND shares a similar direction, but we do not intend to simulate real-world social relationships like DELL, which generally only reflects active users' reactions. Instead, GenFEND generates comments from multiple subpopulations within diverse views and makes veracity judgments based on a comprehensive understanding and analysis of both active and silent users.

\section{Proposed Framework: GenFEND}
Fake news detection is generally formulated as a binary classification task between fake and real news~\cite{Shu_FakeNewsSurvey}.
Given a news piece $o$ (and elicited comments $C_{actual}$), we first generate a group of comments $C$ based on $o$. 
With the learned news piece's feature $\boldsymbol{e}^o$ (actual comments' feature $\boldsymbol{e}^c_{actual}$) and generated comments' feature $\boldsymbol{r}$, 
we aim to learn a content-only model $f(\boldsymbol{e}^o, \boldsymbol{r}) \rightarrow y$ or a comment-based model $f(\boldsymbol{e}^o, \boldsymbol{e}^c_{actual}, \boldsymbol{r}) \rightarrow y$, where $y$ is the ground-truth veracity label. 
\figurename~\ref{fig:model} overviews the proposed framework, GenFEND, which consists of multi-view comment generation, multi-subpopulation feedback understanding, and aggregation and classification. 
\subsection{Multi-View Comment Generation}
\label{sec: generation}
Given a news piece $o$, we aim to generate diverse comments from different types of users.
We first select three typical user attributes, i.e., gender, age, and education, which closely correlate with the difference between real and fake news.
Existing studies reveal that there are gender differences of interest in discussion participation for different fake news topics~\cite{Gender_FakeNews, ipm}, reflecting their different viewpoints. Controlling the gender attribute can capture such differences for comment generation.
Moreover, age and educational level are empirically proved to correlate with cognitive abilities such as remembering and understanding, significantly influencing belief in fake news~\cite{Gaillard_Cognitive}. Including the two factors could better cover the target audiences of specific fake news.
These attributes can certainly be from one single user, so we utilize all three demographic attributes mentioned above to set user profiles from multiple views via combination\footnote{Due to the limit of generation cost, we did not cover all possible user attributes but chose the three frequently considered ones based on existing studies.}.
Specifically, the assignments to the three attributes are:
\begin{itemize}[nosep,leftmargin=1em,labelwidth=*,align=left]
    \item \textbf{Gender:} male; female.
    \item \textbf{Age:} under 17 years old; 18 to 29 years old; 30 to 49 years old; 50 to 64 years old; over 65 years old.
    \item \textbf{Education:} a college graduate; has not graduated from college; has a high school diploma or less.
\end{itemize}

By combining different assignments of gender, age, and education, we obtain 30 different user profiles.
The combinations are then used to prompt the LLM to generate comments by role-playing the 30 types of users:

\begin{tcolorbox}[title=Prompt 1: Comment Generation Prompt, boxrule=0pt, left=1mm, right=1mm, top=1mm, bottom=1mm, fontupper=\small]
    \textbf{System Prompt:} Suppose you are a [\emph{gender}] Twitter user. You are [\emph{age}] . Educationally, you [\emph{education}]. You will be provided with an article.
    You should write one comment about the article. Note that your comment needs to match your identity, and should be brief and natural, like normal Twitter users. \\
    \textbf{Context Prompt:} news: [\emph{the given news $o$}]
\end{tcolorbox}

\subsection{Multi-Subpopulation Feedback Understanding}
To comprehensively analyze user feedback produced in \S~\ref{sec: generation}, we propose understanding these comments from a multi-subpopulation perspective, where the subpopulation groups are split based on the attributes in the three views. For example, there are two groups from the gender view, male and female.
Given the news piece $o$ and generated $n$ corresponding comments $C = \{c_1, ..., c_n\}$, we adopt pre-trained sentence transformers to encode comments $C$ into $dim$-dimensional embeddings as $E^c = \{\boldsymbol{e}^c_1, ..., \boldsymbol{e}^c_n\}$, where $\boldsymbol{e}^c_i \in \mathbb{R}^{dim}, 1 \leq i \leq n$.
We denote gender, age, and education views as $\mathcal{G}, \mathcal{A}, \mathcal{E}$ respectively.
For each view $\mathcal{V} \in \{\mathcal{G}, \mathcal{A}, \mathcal{E}\}$, we split user comments $C$ into $m_\mathcal{V}$ different subpopulation groups, which are $C_1, ...,C_{m_\mathcal{V}}$, where $C_1 \cup ... \cup C_{m_\mathcal{V}} = C$, and $C_p \cap C_q = \emptyset (p \neq q$ and $p, q \in \{1, ..., m_\mathcal{V}\})$.
Here, $m_\mathcal{G} = 2, m_\mathcal{A} = 5, \text{and } m_\mathcal{E} = 3$. 
The corresponding comments embeddings in subpopulation group $C_p$ are denoted as $E^{c_p} = \{\boldsymbol{e}^c_i\}_{c_i \in C_p}$.
To perceive the reactions reflected by comments from different groups, we adopt two operations to reflect the overall viewpoint shared among users in the same group and the discrepancy across different groups respectively.

\textbf{Overall Semantic Feature Extraction.} To understand users' overall viewpoints in the same subpopulation group, we average embeddings of comments for each subpopulation group. The semantic feature for the subpopulation group $C_p$ in view $\mathcal{V}$ is formulated as:
\begin{align}
    \boldsymbol{s}^{mean}_p = \frac{1}{|E^{c_p}|}\sum_{\boldsymbol{e}^c_i \in E^{c_p}}{\boldsymbol{e}^c_i},
\end{align}
where $p \in \left\{1, ..., m_\mathcal{V}\right\}$, and $|E^{c_p}|$ is the number of comments in subpopulation group $C_p$.

\textbf{Diversity Representation Extraction.} The divergence of viewpoints among different subpopulation groups indicates the diversity of feedback under the specific view.
We assume different dimensions distributed in the comments' embedding space represent different viewpoints and we calculate the KL divergence based on the comments' embedding to measure the diversity of viewpoints for each view.
Specifically, for each pair of subpopulation groups $C_p$ and $C_q$, the distribution divergence is calculated as follows:
\begin{equation}
     d_{p, q}
    = \frac{1}{|E^{c_p}||E^{c_q}|}\sum_{\boldsymbol{e}^c_i \in E^{c_p}, e^c_j \in E^{c_q}}\text{kl\_div}(\hat{e}^c_i, \hat{e}^c_j),
\end{equation}
where $E^{c_p}$ and $E^{c_q}$ are embeddings of comments in the subpopulation group $C_p$ and $C_q$ respectively, $|E^{c_p}|$ and $|E^{c_q}|$ are the numbers of comments in $C_p$ and $C_q$.
$\hat{e}^c_i = \text{Softmax}(e^c_i)$, which transform the embeddings into probability distributions.
$\text{kl\_div}(\cdot, \cdot)$ is the Kullback-Leiber divergence operation.
After calculating the distribution divergence between every pair of subpopulation groups, we obtain the diversity representation $\boldsymbol{d}^\mathcal{V}$ for view $\mathcal{V}$ as follows:
\begin{align}
    \boldsymbol{d}^\mathcal{V} = \bigoplus_{p, q \in \left\{1,... ,m_\mathcal{V}\right\}, p \neq q} d_{p, q}.
\end{align}

Finally, we obtain semantic features $\boldsymbol{s}^{mean}_p$ for each subpopulation group $C_p$ and diversity representation $\boldsymbol{d}^\mathcal{V}$ for each view $\mathcal{V} \in \{\mathcal{G}, \mathcal{A}, \mathcal{E}\}$.

\subsection{Aggregation and Classification}
To better enhance fake news detection performance, it is necessary to generate high-quality comment representations that can represent user feedback from multiple subpopulation groups.
Note that different views specialize in different analyzing perspectives, and comments from different subpopulation groups in each view reflect the characteristics of viewpoints from such perspectives.
Therefore, to aggregate comments features of multi-subpopulation groups, both intra-view aggregation and inter-view aggregation are considered.

\textbf{Intra-View Aggregation.} Cross-subpopulation correlations can provide complementary information at each view.
Therefore, we design a cross-subpopulation fusion module to learn the overall semantic feature for each view $\mathcal{V}$.

Specifically, given the news' content feature $\boldsymbol{e}^o \in \mathbb{R}^{dim}$ and the comments' semantic feature $\boldsymbol{s}^{mean}_p$ of each subpopulation group $C_p$, we obtain weights $w^\mathcal{V}$ for subpopulation groups by calculating the dot-product between $\boldsymbol{e}^o$ and $\left\{\boldsymbol{s}^{mean}_p\right\}_{p \in \left\{1, ..., m_\mathcal{V}\right\}}$, which can be formulated as:
\begin{align}
\boldsymbol{s}^\mathcal{V}_{cat} &= \left[ \boldsymbol{s}^{mean}_1; ...; \boldsymbol{s}^{mean}_{m_\mathcal{V}}\right], \\
w^\mathcal{V} &= \text{Softmax}(\boldsymbol{s}^\mathcal{V}_{cat} \cdot {\boldsymbol{e}^o}^\mathsf{T} / \sqrt{dim}),
\end{align}
where $\boldsymbol{s}^\mathcal{V}_{cat} \in \mathbb{R}^{m_\mathcal{V} \times dim}$ denotes the stacking of semantic features of all subpopulation groups in the view $\mathcal{V}$,
$w^\mathcal{V} \in \mathbb{R}^{m_\mathcal{V} \times 1}$ denotes the weights for subpopulation groups,
and $dim$ is the embedding dimension of comments features.

With $w^\mathcal{V}$, we operate subpopulation-level aggregation to obtain the semantic feature $\boldsymbol{s}^\mathcal{V} \in \mathbb{R}^{dim}$ for view $\mathcal{V}$ as follows:
\begin{equation}
    \boldsymbol{s}^{\mathcal{V}} = {w^\mathcal{V}}^\mathsf{T} \cdot \boldsymbol{s}^\mathcal{V}_{cat}.
\end{equation}

\textbf{Inter-View Aggregation.} The subpopulation-level divergences and the topic of the news content can help measure the relative importance of different views.
To this end, we utilize a view gate to aggregate the three views adaptively.
We use the news content feature $\boldsymbol{e}^o \in \mathbb{R}^{dim}$ and diversity representation $\boldsymbol{d} = \oplus_{\mathcal{V} \in \left\{\mathcal{A}, \mathcal{G}, \mathcal{E}\right\}}\boldsymbol{d}^{\mathcal{V}}$ as input to guide the aggregation.
The inter-view aggregation module outputs a vector $\boldsymbol{a}$, denoting the weight of each view for a specific news piece:
\begin{equation}
    \boldsymbol{a} = \text{Softmax}(G(\boldsymbol{e}^o \oplus \boldsymbol{d} ; \theta)),
\end{equation}
where $G(\cdot ; \theta)$ is the view gate, $\theta$ is the parameters of the view gate, and the view gate $G(\cdot ; \theta)$ is a two-layer feed-forward network.
We use $\text{Softmax}(\cdot) $ to normalize the output of $G(\cdot;\theta)$ and $\boldsymbol{a} = \left[a^\mathcal{G}, a^\mathcal{A}, a^\mathcal{E}\right]$ is the weight vector denoting the importance of each view.
The generated comments' final feature is:
\begin{equation}
    \boldsymbol{r} = \sum_{\mathcal{V} \in \left\{\mathcal{G}, \mathcal{A}, \mathcal{E}\right\}} a^\mathcal{V}\boldsymbol{s}^\mathcal{V}.
\end{equation}

\textbf{Classification.} With the aggregated representation $\boldsymbol{r}$, we predict the probability of news piece $o$ being fake with:
\begin{equation}
    \hat{y} = \begin{cases}
        \text{Sigmoid}(\text{MLP}(\boldsymbol{r} \oplus \boldsymbol{e}^o)) & \text{\emph{w/o} actual cmts}, \\
        \text{Sigmoid}(\text{MLP}(\boldsymbol{r} \oplus \boldsymbol{e}^o \oplus \boldsymbol{e}^c_{actual})) & \text{\emph{w/} actual cmts}.
    \end{cases}
\end{equation}
We optimize all the parameters by minimizing the cross-entropy loss with backpropagation. The loss function for one sample is:
\begin{equation}
    \mathcal{L} = -y\log\hat{y} - (1-y)\log(1-\hat{y}),
\end{equation}
where $y$ is the ground-truth label of news piece $o$ (1 for fake and 0 for real), and  $\hat{y} \in [0,1] $ is the predicted probability of $o$ being fake.
\begin{table}[htbp]
  \centering
  \small
  \caption{\label{tab:dataset} Datasets statistics.}
    \setlength{\tabcolsep}{4pt}
    \begin{tabular}{lrrrrrrr}
    \toprule
    \multicolumn{1}{c}{\multirow{2}[4]{*}[0.3em]{\textbf{Dataset}}} & \multicolumn{2}{c}{\multirow{1}[1]{*}[0.2em]{\textbf{Train}}} & \multicolumn{2}{c}{\multirow{1}[2]{*}[0.3em]{\textbf{Validation}}} & \multicolumn{2}{c}{\textbf{Test}} & \multicolumn{1}{c}{\multirow{2}[4]{*}[0.1em]{\textbf{Total}}} \\
\cmidrule{2-7}          & \multicolumn{1}{l}{Fake} & \multicolumn{1}{l}{Real} & \multicolumn{1}{l}{Fake} & \multicolumn{1}{l}{Real} & \multicolumn{1}{l}{Fake} & \multicolumn{1}{l}{Real} \\
    \midrule
    \textbf{Weibo21} & 2,883 & 2,179 & 540 & 702 & 539 & 724 & 7,567\\
    \textit{\#comments} & \textit{76,015} & \textit{72,152} & \textit{8,234} & \textit{9,766} & \textit{8,576} & \textit{10,032} & \textit{184,775}\\
    \textbf{GossipCop} & 1,816 & 3,775 & 552 & 820 & 844 & 483 & 8,260\\
    \textit{\#comments} & \textit{16,823} & \textit{29,723} & \textit{4,054} & \textit{6,046} & \textit{3,652} & \textit{6,091} & \textit{66,389}\\
    \textbf{LLM-mis} & 410 & 289 & 118 & 82 & 59 & 42 & 1,000\\
    \textit{\#comments} & \multicolumn{7}{c}{\textit{(no comments provided)}}\\
    \bottomrule
    \end{tabular}%
  \vspace{-0.4cm}
\end{table}%

\section{Experiments}
The experiments are to answer the following evaluation questions:
\begin{itemize}[nosep,leftmargin=1em,labelwidth=*,align=left]
    \item[\textbf{EQ1} ]Can GenFEND improve fake news detection performance?
    \item[\textbf{EQ2} ]How effective is the GenFEND archtecture?
    \item[\textbf{EQ3} ]How effective are generated comments and why?
\end{itemize}

\subsection{Experimental Setup}
\textbf{Datasets.} 
We conduct experiments on three public datasets, including human-written \textbf{Weibo21}~\cite{Nan_MDFEND}, \textbf{GossipCop}~\cite{Shu_FakeNewsNet}, and the LLM-generated \textbf{LLM-mis}~\cite{Chen_LLM_Mis}.
For the Weibo21 and GossipCop datasets, we did the train-validation-test set split chronologically to simulate real-world scenarios. For LLM-mis, we did a random split because of the lack of actual timestamps for generated samples.
\tablename~\ref{tab:dataset} shows the dataset statistics.

\textbf{Baselines.} 
For Weibo21 and GossipCop datasets, we compare with two groups of fake news detection methods.
The first group is content-only methods:
(1) \textbf{LLM \emph{w/} cnt}: A zero-shot method that directly prompts an LLM to make veracity judgments with only news content provided; 
(2) \textbf{BERT}~\cite{devlin_BERT}: A pre-trained language model that is widely used as the text encoder for fake news detection~\cite{wu2020adaptive,fakebert,Nan_MDFEND}, with the last layer finetuned conventionally; 
(3) \textbf{ENDEF}~\cite{Zhu_ENDEF}: A method that removes entity bias to obtain generalizable features; 
(4) \textbf{EANN-text}~\cite{Wang_EANN}: A model which aims to learn event-invariant representations for fake news detection. 
Here we use its text-only version.
The second group is comment-based methods: 
(1) \textbf{LLM \emph{w/} actual cmts}: A zero-shot method that directly prompts an LLM to make veracity judgments with both news content and actual comments provided; 
(2) \textbf{dEFEND}~\cite{Shu_dEFEND}: A model that develops a sentence-comment co-attention sub-network for fake news detection; 
(3) \textbf{DualEmo}~\cite{Zhang_DualEmo}: A framework that considers both publisher emotion, social emotion, and their gap for fake news detection; 
(4) \textbf{CAS-FEND(tea)}~\cite{Nan_CASFEND}: The CAS-FEND teacher module that exploits user comments from both semantic and emotional aspects.

For the LLM-mis dataset which only contains generated misinformation samples, we include LLM \emph{w/} cnt, BERT~\cite{devlin_BERT}, ENDEF~\cite{Zhu_ENDEF}, and \textbf{DELL}~\cite{Wan_DELL} for comparison. DELL is a recently proposed method that exploits LLMs to generate user comments and perform proxy tasks for fake news detection.
It uses four ensemble strategies to get the final judgment.
In our experiments, we use the same user attributes for DELL and GenFEND to ensure fairness.

\begin{table*}[htbp]
  \centering
  \small
  \caption{Performance of LLM prediction (\emph{w/} cnt and \emph{w/} actual cmts), and other content-only and comment-based methods with or without GenFEND on the Weibo21 and GossipCop datasets. The better result for each comparison pair is \textbf{bolded}.}
    
    \begin{tabular}{llrrrrrrrrrr}
    \toprule
    \multicolumn{1}{c}{\multirow{2}[4]{*}[0.3em]{\textbf{Category}}}
    & \multicolumn{1}{c}{\multirow{2}[4]{*}[0.3em]{\textbf{Method}}} & \multicolumn{5}{c}{\textbf{Weibo21}}           & \multicolumn{5}{c}{\textbf{GossipCop}} \\
\cmidrule{3-12}      &    & \multicolumn{1}{c}{macF1} & \multicolumn{1}{c}{Acc} & \multicolumn{1}{c}{AUC} & \multicolumn{1}{c}{F1-real} & \multicolumn{1}{c}{F1-fake} & \multicolumn{1}{c}{macF1} & \multicolumn{1}{c}{Acc} & \multicolumn{1}{c}{AUC} & \multicolumn{1}{c}{F1-real} & \multicolumn{1}{c}{F1-fake} \\
    \midrule
    \multicolumn{1}{c}{\multirow{7}[2]{*}{\textbf{Cnt-Only Methods}}} 
    & LLM \emph{w/} cnt & 0.6795 & 0.6825 & 0.7119 & 0.6486 & 0.7105 & 0.6029 & 0.6774 & 0.6043 & 0.7750 & 0.4309 \vspace{0.2mm}\\
    \cdashline{2-12} \vspace{-3mm} \\
    & BERT  & 0.7625  & 0.7633  & 0.8439  & 0.7749  & 0.7500  & 0.8073  & 0.8259  & 0.8931  & 0.8670  & 0.7477  \\
    & \quad \emph{w/} GenFEND & \textbf{0.7926} & \textbf{0.7935} & \textbf{0.8648} & \textbf{0.8079} & \textbf{0.7769} & \textbf{0.8457} & \textbf{0.8576} & \textbf{0.9137} & \textbf{0.8885} & \textbf{0.8029} \\
    & ENDEF & 0.7701  & 0.7717  & 0.8477  & 0.7870  & 0.7532  & 0.8298  & 0.8463  & 0.9002  & 0.8826  & 0.7770  \\
    & \quad \emph{w/} GenFEND & \textbf{0.7898} & \textbf{0.7900} & \textbf{0.8617} & \textbf{0.7923} & \textbf{0.7775} & \textbf{0.8395} & \textbf{0.8515} & \textbf{0.9131} & \textbf{0.8835} & \textbf{0.7954} \\
    & EANN-text  & 0.7212  & 0.7240  & 0.7986  & 0.7467 & 0.6956  & 0.8179  & 0.8348  & 0.8904  & 0.8733  & 0.7626  \\
    & \quad \emph{w/} GenFEND & \textbf{0.7497} & \textbf{0.7560} & \textbf{0.8100} & \textbf{0.7603} & \textbf{0.7273} & \textbf{0.8279} & \textbf{0.8425} & \textbf{0.8969} & \textbf{0.8780} & \textbf{0.7779} \\
    \midrule
    \multicolumn{1}{c}{\multirow{6}[4]{*}{\textbf{Cmt-Based Methods}}} 
    & LLM \emph{w/} actual cmts & 0.7663 & 0.7664 & 0.7868 & 0.7607 & 0.7718 & 0.6360 & 0.6654 & 0.6351 & 0.7394 & 0.5326 \vspace{0.2mm}\\
    \cdashline{2-12} \vspace{-3mm} \\
    & dEFEND & 0.7995 & 0.8005 & 0.8832 & 0.8133 & 0.7857 & 0.8670  & 0.8794  & 0.9382  & 0.9076  & 0.8265  \\
    & \quad \emph{w/} GenFEND & \textbf{0.8102} & \textbf{0.8188} & \textbf{0.8875} & \textbf{0.8295} & \textbf{0.7991} & \textbf{0.8904} & \textbf{0.8913} & \textbf{0.9581} & \textbf{0.9131} & \textbf{0.8512} \\
    
    & DualEmo & 0.7834  & 0.7837  & 0.8823  & 0.7987  & 0.7925  & 0.8864  & 0.8802  & 0.9341  & 0.9040  & 0.8620  \\
    & \quad \emph{w/} GenFEND & \textbf{0.8083} & \textbf{0.8084} & \textbf{0.8992} & \textbf{0.8120} & \textbf{0.8102} & \textbf{0.9004} & \textbf{0.9135} & \textbf{0.9557} & \textbf{0.9358} & \textbf{0.8688} \\
    
    & CAS-FEND(tea) & 0.8181  & 0.8187  & 0.9016  & 0.8287  & 0.8074  & 0.9188  & 0.9261  & 0.9716  & 0.9432  & 0.8944  \\
    & \quad \emph{w/} GenFEND & \textbf{0.8217} & \textbf{0.8200} & \textbf{0.9094} & \textbf{0.8309} & \textbf{0.8112} & \textbf{0.9250} & \textbf{0.9398} & \textbf{0.9822} & \textbf{0.9477} & \textbf{0.9084} \\
    \bottomrule
    \end{tabular}%
  \label{tab:main_experiment}%
\end{table*}%

\textbf{Implementation Details.} 
We prompt GLM-4~\cite{glm-4} for Weibo21, and GPT-3.5-Turbo (\textit{version 0125})~\cite{chatgpt} for GossipCop in comment generation and veracity judgment.
For comment generation, the sampling temperature is set to 0.95 for GLM-4 and 1.0 for GPT-3.5-Turbo to guarantee diversity and creativity. \emph{max\_tokens} is 100.
For veracity judgment, the sampling temperature is set to 0.1 for GLM-4 and GPT-3.5-Turbo to get definitive answers (see Prompt 2, in which underlined text is only for \textbf{LLM \emph{w/} actual cmts}).
For non-LLM methods, we adopt \emph{bert-base-chinese} for Weibo21 and \emph{bert-base-uncased} for GossipCop/LLM-mis to encode news content. 
The maximum number of tokens of news content is 170.
We adopt sentence transformers to obtain comment embeddings (\emph{Dmeta-embedding-zh}\footnote{https://huggingface.co/DMetaSoul/Dmeta-embedding-zh} for Weibo21 and \emph{bge-large-en-v1.5}\footnote{https://huggingface.co/BAAI/bge-large-en-v1.5} for GossipCop and LLM-mis). $dim$ is set to 768 and 1024 for Chinese and English, respectively. 
To evaluate the effect of GenFEND, we utilize the feature of news content from the baseline methods as $\boldsymbol{e}^o$ in \figurename~\ref{fig:model}.
For comment-based methods, we concatenate $\boldsymbol{e}^o$ with the feature of actual comments for final classification.
\begin{tcolorbox}[title=Prompt 2:  Prompt for Veracity Judgment, boxrule=0pt, left=1mm, right=1mm, top=1mm, bottom=1mm, fontupper=\small]
    \textbf{System Prompt:} Given the following news piece \dashuline{and the corresponding comments}, predict the veracity of this news piece. \dashuline{The comments are collected from social media users.} If the news piece is more likely to be fake, return 1; otherwise, return 0. Please refrain from providing ambiguous assessments such as undetermined. \\
    \textbf{Context Prompt:} news: [\emph{the given news $o$}]; \dashuline{comments: [\emph{user comments ${c_1, c_2, ...}$}]}. The answer (Arabic numerals) is:
\end{tcolorbox}
\textbf{Metrics.} We report experimental results in five metrics, including the accuracy (Acc), area under the ROC curve (AUC), macro F1 score (macF1), and F1 score for the fake/real class (F1-fake/F1-real).

\subsection{Main Results (EQ1)}
Tables~\ref{tab:main_experiment} and~\ref{tab:LLM_mis} show the results of GenFEND and the compared methods on the three datasets. We have the following observations: 

\begin{table}[htbp]
  \centering
  \small
  \caption{\label{tab:LLM_mis} Performance of LLM \emph{w/} cnt by zero-shot prompting, DELL with different ensemble strategy, and content-only methods (i.e., BERT and ENDEF) with and without GenFEND on the LLM-mis dataset. The best and the second-best results are \textbf{bolded} and \underline{underlined}.}
    \begin{tabular}{lrrrrr}
    \toprule
    \multicolumn{1}{c}{\textbf{Method}} 
    & \multicolumn{1}{c}{macF1} & \multicolumn{1}{c}{Acc} & \multicolumn{1}{c}{AUC} & \multicolumn{1}{c}{F1-real} & \multicolumn{1}{c}{F1-fake} \\
    \midrule
    \multicolumn{1}{l}{LLM \emph{w/} cnt} & 0.5037 & 0.5050 & 0.5282 & 0.5283 & 0.4792 \\
    \midrule
    DELL \emph{Single}  & 0.8648 & 0.8713 & \underline{0.9503} & 0.8354 & 0.8943 \\
    DELL \emph{Vanilla} & 0.8670 & 0.8713 & 0.9500 & 0.8434 & 0.8908 \\
    DELL \emph{Confidence} & 0.8589 & 0.8614 & 0.9435 & 0.8409 & 0.8772 \\
    DELL \emph{Selective} & 0.8440 & 0.8515 & 0.9402 & 0.8101 & 0.8780 \\
    \midrule
    BERT  &  0.8570 & 0.8614 & 0.9463 & 0.8325 & 0.8816 \\
    \quad \emph{w/} GenFEND & \underline{0.8798} & \underline{0.8837} & \textbf{0.9507} & \underline{0.8588} & \underline{0.9009} \\
    ENDEF & 0.8591 & 0.8614 & 0.9435 & 0.8409 & 0.8772 \\
    \quad \emph{w/} GenFEND & \textbf{0.8883} & \textbf{0.8911} & 0.9435 & \textbf{0.8706} & \textbf{0.9060} \\
    \bottomrule
    \end{tabular}%
    \vspace{-0.4cm}
\end{table}
(1) \textbf{GenFEND brings valuable information beyond news content.}
From the results in the content-only methods (Group 1 in Table~\ref{tab:main_experiment} and Table~\ref{tab:LLM_mis}), we find that content-only methods have improved with GenFEND, even comparable with some comment-based detectors. The results indicate that generated comments are a supplement to news content and a surrogate of actual comments, which can improve early detection performance.

(2) \textbf{GenFEND provides additional information beyond actual comments.}
From the results in the comment-based methods (Group 2 in Table~\ref{tab:main_experiment}), we find that comment-based detectors have improved with GenFEND, which demonstrates a more comprehensive understanding of news content obtained from diverse generated comments.

(3) \textbf{GenFEND performs better than DELL.}
\tablename~\ref{tab:LLM_mis} shows that the best and the second best results in almost all metrics are from BERT \emph{w/} GenFEND and ENDEF \emph{w/} GenFEND, which exceed the performance of DELL with different ensemble strategies.
The superiority of GenFEND indicates the effectiveness of diverse comment generation and subsequent feedback understanding.

(4) \textbf{GenFEND is more effective for identifying fake news than excluding real news.}
In most cases on the weibo21 and GossipCop datasets, GenFEND brings a larger performance improvement in F1-fake than in F1-real, which indicates that the generated comments might contain knowledge and experience that can identify the typical fake news patterns in reality. This makes GenFEND more practical because identifying more fake news pieces is generally a priority in a real-world deployment.
\begin{table}[t]
  \centering
  \small
  \caption{Performance comparison between GenFEND and its ablative variants. ``-'' indicates ``not applicable'', as LLM-mis does not provide actual comments.}
    \begin{adjustbox}{scale = 0.9}
    \begin{tabular}{@{}lrrrrrr@{}}
    \toprule
    \multicolumn{1}{c}{\multirow{2}[4]{*}[0.3em]{\textbf{Method}}} & \multicolumn{2}{c}{\textbf{Weibo21}} & \multicolumn{2}{c}{\textbf{GossipCop}} & \multicolumn{2}{c}{\textbf{LLM-mis}}\\
\cmidrule{2-7}          & \multicolumn{1}{c}{macF1} & \multicolumn{1}{c}{Acc} & \multicolumn{1}{c}{macF1} & \multicolumn{1}{c}{Acc} & \multicolumn{1}{c}{macF1} & \multicolumn{1}{c}{Acc}\\
    \midrule
    BERT \emph{w/} GenFEND & \textbf{0.7926} & \textbf{0.7935} & \textbf{0.8457} & \textbf{0.8576} & 0.8798 & 0.8837 \\
    \quad \emph{w/o} gender view & 0.7768 & 0.7787 & 0.8295 & 0.8448 & 0.8632 & 0.8701 \\
    \quad \emph{w/o} age view & 0.7861 & 0.7870 & 0.8307 & 0.8470 & 0.8699 & 0.8754\\
    \quad \emph{w/o} education view & 0.7836 & 0.7838 & 0.8319 & 0.8515 & 0.8668 & 0.8698 \\
    \quad \emph{w/o} reweighting & 0.7735 & 0.7736 & 0.8320 & 0.8441 & 0.8588 & 0.8679 \\
    \quad \emph{w/o} fusion gate & 0.7897 & 0.7895 & 0.8399 & 0.8501 & 0.8604 & 0.8713\\
    \midrule 
    dEFEND \emph{w/} GenFEND & \textbf{0.8102} & \textbf{0.8188} & \textbf{0.8904} & \textbf{0.8913} &  - & -\\
    \quad \emph{w/o} gender view & 0.8071 & 0.8084 &   0.8697 & 0.8817 & - & -\\
    \quad \emph{w/o} age view &  0.7980 & 0.7989 & 0.8883 & 0.8968 & - & -\\
    \quad \emph{w/o} education view & 0.7988 & 0.7997 & 0.8770 & 0.8892 & - & -\\
    \quad \emph{w/o} reweighting & 0.7954 & 0.7963 & 0.8702 & 0.8813 & - & - \\
    \quad \emph{w/o} fusion gate & 0.8001 & 0.8014 & 0.8787 & 0.8890 & - & - \\
    \bottomrule
    \end{tabular}%
  \label{tab:ablation}%
  \end{adjustbox}
  \vspace{-0.4cm}
\end{table}%

\subsection{Effectiveness of GenFEND Arch. (EQ2)}
We compare with several variants of GenFEND (\S~\ref{sec: ablation}) and decrease the number of actual comments (\S~\ref{sec: actual comments number})
to evaluate the importance of different components in GenFEND and the robustness of GenFEND.
\subsubsection{Importance of Different Views and Their Aggregation}
\label{sec: ablation}
To demonstrate the effectiveness of selected views, we obtain three one-view-removed variants for BERT \emph{w/} GenFEND and dEFEND \emph{w/} GenFEND. As shown in \tablename~\ref{tab:ablation}, the removal of each view consistently caused performance decreases, which indicates that the demographic information from all three views can provide useful information to improve the comments' comprehensiveness and benefit the subsequent analysis in Multi-Subpopulation Feedback Understanding.
As further demonstrated by the case in \tableautorefname~\ref{tab:ablation_case}, the introduction of the gender view could exactly reflect the controversial point concerning the male-female relationship in this fake news piece. After analysis of generated comments from the three views, our proposed GenFEND successfully captured such a key point, assigned the highest weight to the gender view, and correctly identified news fakeness.

To demonstrate the effectiveness of intra- and inter-view aggregations, we replace the \emph{Subpopulation Reweighting} and \emph{Inter-View Fusion Gate} modules with a simple average pooling. From Table~\ref{tab:ablation_case}, we see that the replacement of intra- or inter-view aggregations leads to a performance drop, with a larger drop for the intra-view, which demonstrates that adaptively weighting different subpopulation groups and fusing different views are important for a comprehensive and effective understanding of news. 

\begin{table}[]
    \centering
    \small
     \caption{\label{tab:ablation_case} A fake news case about a family affair, where the gender view gains the highest weight.}
    \begin{tabular}{@{}ccc@{}}
    \toprule
         \multicolumn{3}{p{0.45\textwidth}}{\textbf{News:} According to a news report, a man consecutively heard phone rings at 02:01 am on November 11 and noticed many weird transactions of his four credit cards! He initially thought his cards had been stolen, but upon closer inspection, he found that it was all from online shopping malls. When he went to the study room, he found his wife shopping on the computer. After figuring out that about 230,000 CNY was spent in total, he suddenly fainted. His wife dialed 120. Despite an attempt at resuscitation, he finally died at 6 am.} \\
        \midrule
         \multicolumn{1}{l}{\textbf{gender view: } \textbf{\emph{0.5442}} } &
         \multicolumn{1}{l}{\textbf{age view: } \emph{0.1459}} &
         \multicolumn{1}{l}{\textbf{education view: } \emph{0.3110}} \\
    \bottomrule
    \end{tabular}
    \vspace{-0.2cm}
\end{table}

\subsubsection{Robustness of GenFEND when Decreasing the Number of Actual Comments}
\label{sec: actual comments number}
To explore GenFEND's capability of early detection, we evaluate the impact of GenFEND for comment-based methods when the number of actual comments varies.
Experimental results from previous sections demonstrate the effectiveness of generated comments partnering with full actual comments
It is more inspiring to investigate whether GenFEND can improve the fake news detection performance of comment-based methods with fewer actual comments available (i.e., at the earlier stage).

\begin{figure}
    \centering
    \includegraphics[width = 1.0\linewidth]{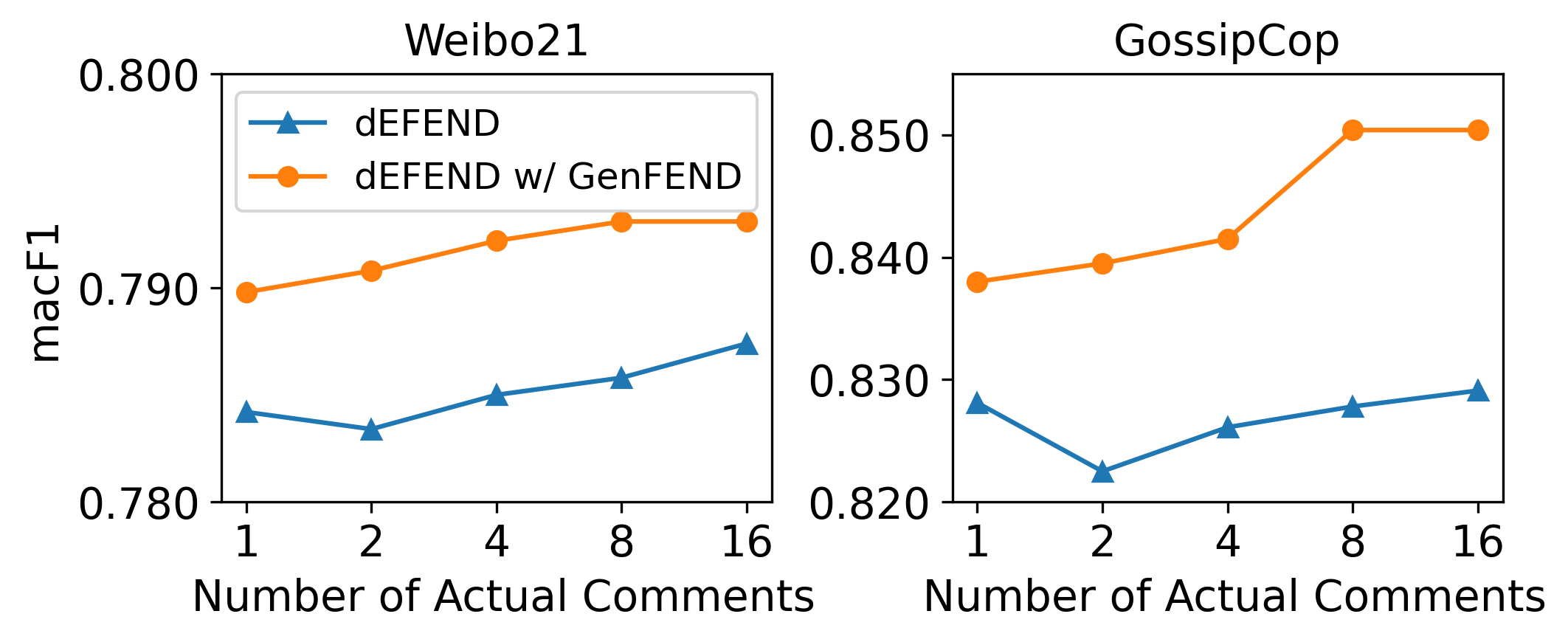}
    \caption{Early detection performance of dEFEND and dEFEND \emph{w/} GenFEND with 1, 2, 4, 8, and 16 actual comments for the testing data.}
    \label{fig:early}
    \vspace{-0.5cm}
\end{figure}

We experiment with dEFEND and set the number of actual comments of testing data to 1, 2, 4, 8, and 16, and compare the two versions with or without GenFEND.
As depicted by \figureautorefname~\ref{fig:early}, dEFEND \emph{w/} GenFEND surpasses dEFEND when the number of actual comments is limited, which indicates that GenFEND improves dEFEND's early detection performance.

\begin{table*}[t]
  \centering
  \small
  \caption{Performance comparison with actual comments and generated comments for LLM, content-only method BERT and comment-based method dEFEND. The best results for each comparison pair are \textbf{bolded}.}
    \begin{tabular}{rcrrrrrrrrrr}
    \toprule
\multicolumn{1}{c}{\multirow{2}[4]{*}[0.3em]{\textbf{Method}}} & \multicolumn{1}{c}{\multirow{2}[4]{*}[0.3em]{\textbf{Comment Type}}} & \multicolumn{5}{c}{\textbf{Weibo21}}           & \multicolumn{5}{c}{\textbf{GossipCop}} \\
\cmidrule{3-12}          &       & \multicolumn{1}{c}{macF1} & \multicolumn{1}{c}{Acc} & \multicolumn{1}{c}{AUC} & \multicolumn{1}{c}{F1-real} & \multicolumn{1}{c}{F1-fake} & \multicolumn{1}{c}{macF1} & \multicolumn{1}{c}{Acc} & \multicolumn{1}{c}{AUC} & \multicolumn{1}{c}{F1-real} & \multicolumn{1}{c}{F1-fake} \\
\midrule
\multicolumn{1}{l}{\multirow{2}[1]{*}{LLM \emph{w/} comment}} &  actual & \textbf{0.7597} & \textbf{0.7601} & \textbf{0.7824} & 0.7506 & \textbf{0.7689} & 0.6360 & 0.6654 & 0.6351 & 0.7394 & 0.5326 \\
\multicolumn{1}{l}{} & generated & 0.7403 & 0.7482 & 0.7384 & \textbf{0.7857} & 0.6984 & \textbf{0.6567} & \textbf{0.6917} & \textbf{0.6532} & \textbf{0.7664} & \textbf{0.5471} \\  
\midrule
\multicolumn{1}{l}{\multirow{2}[1]{*}{BERT \emph{w/} GenFEND}} &  actual & 0.7805 & 0.7816 & 0.8540 & 0.8048 & 0.7762 & 0.8390 & 0.8523 & \textbf{0.9189} & 0.8852 & 0.7928 \\
\multicolumn{1}{l}{} & generated &  \textbf{0.7926} & \textbf{0.7935} & \textbf{0.8648} & \textbf{0.8079} & \textbf{0.7769} & \textbf{0.8457} & \textbf{0.8576} & 0.9137 & \textbf{0.8885} & \textbf{0.8029} \\  
\midrule
\multicolumn{1}{l}{\multirow{2}[1]{*}{dEFEND \emph{w/} GenFEND}} & actual & 0.7995 & 0.8005 & 0.8832 & 0.8133 & 0.7857 & 0.8670  & 0.8794  & 0.9382  & 0.9076  & 0.8265 \\
\multicolumn{1}{l}{} & generated &  \textbf{0.8102} & \textbf{0.8188} & \textbf{0.8875} & \textbf{0.8295} & \textbf{0.7991} & \textbf{0.8904} & \textbf{0.8913} & \textbf{0.9581} & \textbf{0.9131} & \textbf{0.8512} \\
\bottomrule
\end{tabular}%
\label{tab:different comment}%
\end{table*}%

\subsection{Effectiveness of Generated Comments (EQ3)}
To investigate the impact of generated comments, we analyze from three aspects: comparing generated comments with actual comments (\S~\ref{section: comments_comparison}), comparing generated comments from silent and active users (\S~\ref{section: user_comparison}), analyzing the diversity of users (\S~\ref{section: user_diversity}), and evaluating the generated comments' conformity to pre-defined user attributes (\S~\ref{sec: user study}).
\subsubsection{Comparison of Generated Comments and Actual Comments}
\label{section: comments_comparison}
To evaluate the effect of generated comments compared to actual comments, we adopt LLM \emph{w/} comment, BERT \emph{w/} GenFEND, and dEFEND \emph{w/} GenFEND by providing actual comments and generated comments respectively. 
To obtain the user profiles for actual comments, we prompt GPT-3.5-Turbo by calling its API to predict pseudo user profiles:
\begin{tcolorbox}[title= Prompt 3: Prompt for User Profile Prediction, boxrule=0pt, left=1mm, right=1mm, top=1mm, bottom=1mm]
    \textbf{System Prompt:} Given a news-comment pair, you should predict the commenter's gender, age, and education level. Note the gender should be chosen from \{male, female\}; the age should be chosen from \{$\leq$17; 18-29; 30-49; 50-64; $\geq$65\}; the education level should be chosen from \{high school diploma or less; an undergraduate; a college graduate\}. Your prediction should follow the format as \{`gender': g; `age': a; `education level': e\}. \\
    \textbf{Context Prompt:} news: [\emph{the given news $o$}]; comment: [\emph{one user comment $c$}]
\end{tcolorbox}
We count the different number of user profiles of actual comments for each news piece and find that only 7 of 30 for Weibo21 and 3 of 30 for GossipCop are covered on average, which can result in many empty subpopulation groups in Multi-Subpopulation Feedback Understanding.
Under the circumstances, we experiment by adding an empty string for all subpopulation groups.
From the experimental results in \tablename~\ref{tab:different comment}, we find that: (1) Generated comments bring more effects than actual comments in almost all cases for BERT/dEFEND \emph{w/} GenFEND, which benefits from the effective collocation of generated diverse comments and the following understanding procedure; (2) The LLM's performance on GossipCop is better when generated comments rather than actual ones are provided, showing the usefulness of generated comments themselves. However, the case is different for Weibo21. We speculate that it is because patterns in actual comments of Weibo21 are easier to capture and different from those in generated ones.

\begin{table}[t]
  \centering
  \small
  \caption{Performance comparison of LLM, content-only method BERT and comment-based method dEFEND with generated comments from active and silent users. The best and the second-best results for each comparison pair are bolded and underlined, respectively.}
    \begin{tabular}{ccrrrr}
    \toprule
\multicolumn{1}{c}{\multirow{2}[4]{*}[0.3em]{\textbf{Method}}} & \multicolumn{1}{c}{\multirow{2}[4]{*}[0.3em]{\textbf{User Type}}} & \multicolumn{2}{c}{\textbf{Weibo21}}           & \multicolumn{2}{c}{\textbf{GossipCop}} \\
\cmidrule{3-6}          &       & \multicolumn{1}{c}{macF1} & \multicolumn{1}{c}{Acc} & \multicolumn{1}{c}{macF1} & \multicolumn{1}{c}{Acc} \\
\midrule
\multicolumn{1}{p{0.11\textwidth}}{\multirow{3}[1]{*}{LLM \emph{w/} gen cmt}} &  active & 0.7334 & \underline{0.7398} & 0.6459 & 0.6806 \\
\multicolumn{1}{l}{} & silent & \underline{0.7379} & 0.7379 & \textbf{0.6586} & \textbf{0.7076} \\
\multicolumn{1}{l}{} & all & \textbf{0.7403} & \textbf{0.7482} & \underline{0.6567} & \underline{0.6917} \\
\midrule
\multicolumn{1}{p{0.11\textwidth}}{\multirow{3}[1]{*}{BERT \emph{w/} GenFEND}} &  active & 0.7825 & 0.7834 & 0.8393 & 0.8523 \\
\multicolumn{1}{l}{} & silent & \underline{0.7925} & \textbf{0.7937} & \underline{0.8400} & \underline{0.8553} \\  
\multicolumn{1}{l}{} & all & \textbf{0.7926} & \underline{0.7935} & \textbf{0.8457} & \textbf{0.8576} \\
\midrule
\multicolumn{1}{p{0.11\textwidth}}{\multirow{3}[1]{*}{dEFEND \emph{w/} GenFEND}} & active & 0.7921 & 0.7931 & 0.8697 & 0.8817 \\
\multicolumn{1}{l}{} & silent & \underline{0.8019} & \underline{0.8025} & \underline{0.8797} & \textbf{0.8922} \\
\multicolumn{1}{l}{} & all & \textbf{0.8102} & \textbf{0.8188} & \textbf{0.8904} & \underline{0.8913} \\
\bottomrule
\end{tabular}%
\label{tab:different user}%
\end{table}%

\subsubsection{Comparison of Generated Comments from Silent and Active Users}
\label{section: user_comparison}
With the pseudo profiles of actual commentators obtained in \S~\ref{section: comments_comparison}, 
we regard users with the same profiles as actual commentators as ``active users'', and the others as ``silent users''.
To prove the importance of diverse users in GenFEND, 
we investigate how the generated comments from silent users and active users contribute to GenFEND respectively.
By referring to actual commentators' pseudo profiles, we split the generated comments into two clusters: silent and active users' comments and conduct experiments to evaluate their impacts.
Specifically, we adopt BERT \emph{w/} GenFEND and dEFEND \emph{w/} GenFEND for evaluation.
Based on the results shown in \tablename~\ref{tab:different user}, we have the following findings:
\textbf{(1) }Only utilizing active users' comments or silent users' comments leads to a performance drop compared to utilizing all generated ones, which indicates that both silent users and active users have positive and complementary effects.
\textbf{(2) }In most cases, models using generated silent users' comments outperform those using active users' ones, showing the former's superior usefulness. This confirms that considering potential silent users and generating such comments inaccessible in reality is helpful for fake news detection.
\begin{figure}
\centering
\includegraphics[width = 1.0\linewidth]{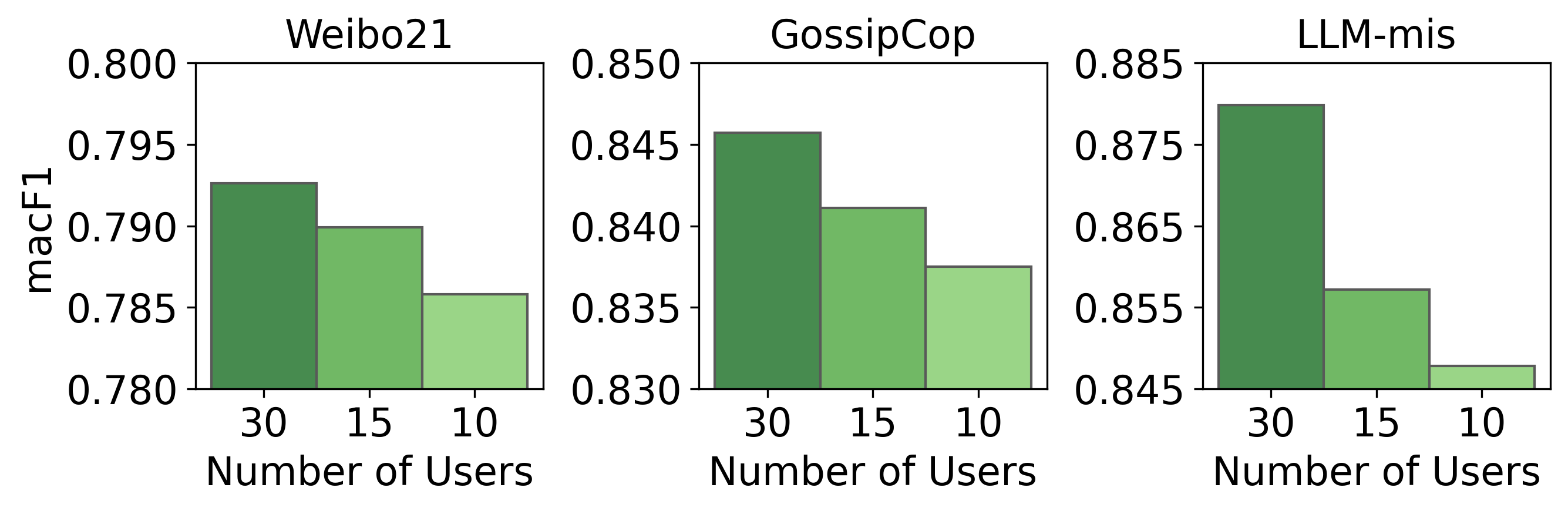}
\caption{Macro F1 scores of BERT \emph{w/} GenFEND with 30 generated comments from different numbers of users (30/15/10). Each user generates the same number of comments.}
\label{fig:diversity}
\vspace{-0.5cm}
\end{figure}

\subsubsection{Impacts of User Diversity}
\label{section: user_diversity}
To conduct a quantitative analysis of the impact of user diversity, we experiment with three groups of generated comments that cover different numbers of user types.
Specifically, besides the generated comments in the main experiment (Table~\ref{tab:main_experiment}), we prompt LLMs to generate another two pieces of comments for each type of user and obtain 90 generated comments in total.
For a fair comparison, we guarantee that the three groups contain the same number of comments in total (here, 30), i.e., 30 different users $\times$ 1 comment per user for Group 1, 15 $\times$ 2 for Group 2, and 10 $\times$ 3 for Group 3.
We use the full set of user types for Group 1, randomly select 15 out of the 30 types for Group 2, and again randomly select 10 out of 15 for Group 3. We re-use the same 30 generated comments in the main experiments for Group 1.

As shown in \figurename~\ref{fig:diversity}, with the three groups of comments provided, the macro F1 score of BERT \emph{w/} GenFEND decreases when the number of users (of different types) decreases, even if more comments are provided by each user.
This confirms our initial assumption that user diversity is of great importance for comment generation in fake news detection.

\begin{figure}
    \centering
    \includegraphics[width = 1.0\linewidth]{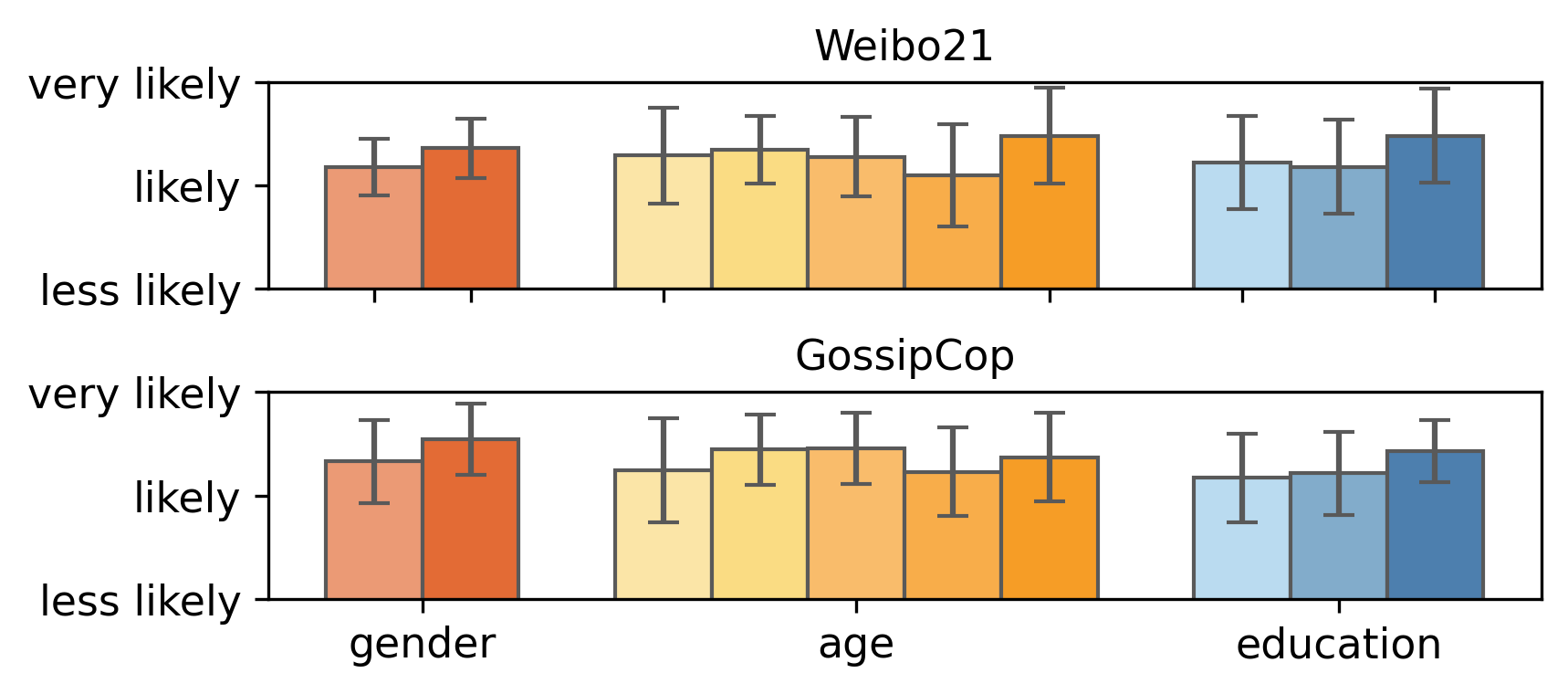}
    \caption{Average conformity scores of generated comments to each attribute. The bars are in ``male; female'' order for gender, from young to old for age, and from low-level to high-level degree for education.}
    \label{fig:conformity}
    \vspace{-0.5cm}
\end{figure}

\subsubsection{Conformity Evaluation of Generated Comments to User Attributes}
\label{sec: user study}
We set a human evaluation to measure how generated comments conform to specified user attributes. 
Specifically, we use a subset of 300 generated comments from Weibo21 and GossipCop (150 for each), which covers all 30 user types.
Given a news piece and its corresponding generated comments, we ask participants to judge how possible a given comment is from the user with pre-defined attributes (less likely/likely/very likely). 
We recruited 20 participants (experienced in social media discussion), each of whom accomplished the task in about one hour on average and was paid their hourly income. 
As depicted in Figure~\ref{fig:conformity}, generated comments are scored with a high degree of conformity to user attributes, maintaining user and comment diversity.

\begin{table*}[htbp]
  \centering
  \small
  \caption{\label{tab:cases} Three cases from the testing set. In Cases 1 and 2, the detector predicted incorrectly using actual comments but did correctly using mixed (actual and generated) comments. Case 3 is a failure case, even if mixed comments were provided. For all cases, we list at most three representative comments in the actual and generated comment list for brevity.}
    \begin{tabular}{p{1.0\textwidth}}
    \toprule
    \textbf{(Case 1) Real News:} Many people believe that yogurt is rich in probiotics, which can balance intestinal flora and promote intestinal motility. Actually, research finds that yogurt does not improve digestion.\newline{}\textbf{Actual Comments:} \ding{182} Well, it is the first time I've heard about it. \ding{183} Really?\newline{}\textbf{Generated Comments:} \ding{182} It turns out that drinking yogurt may not necessarily promote digestion. Knowledge received! I usually like to drink yogurt. Seems that I need to pay more attention to my diet. \ding{183} So that is what it is. I used to believe yogurt helps digestion. It seems that we should pay more attention to food pairing and not just rely on yogurt. \ding{184} I have always thought that yogurt is helpful for digestion. It seems that learning more professional knowledge is needed when selecting foods later. I will forward it to everyone. Don't be misled by wrong knowledge anymore! \\
    \midrule
     \textbf{(Case 2) Fake News:} There is a rumor that the Maserati driver, a woman who caused the car accident, is suffering from mental illness.\newline{}\textbf{Actual Comments:} \ding{182} Money makes the mare go. \ding{183} So simple. She has a father being in power in the government! \ding{184} Every time something goes wrong, they use this excuse. Can't they find some new ones? If she has a mental illness, then let her stay in a mental hospital!\newline{}\textbf{Generated Comments:} \ding{182} It is too outrageous! How could someone use mental illness as an excuse? \ding{183} Is it reliable? If that's the case, the situation will be complicated. Hope the departments concerned can provide accurate inquiry results. \ding{184} Oh god. How could this be? Hope it will come out in the wash ASAP. \\
    \midrule
    \textbf{(Case 3, Failure Case) Fake News:} Recently in Guangzhou. Miss Li, a female foreign teacher from the UK, mainly worked as an onsite tutor for students. Though her course was expensive, she was still very popular among students. However, this foreign teacher's unique skill was to have a sexual relationship with those who had a good English score! A parent discovered this and called the police. Further investigation revealed that five other students had the same experience. Such a ``reward'' is almost unheard of. It is truly despicable for such an expensive foreign language teacher to do such an evil.\newline{}\textbf{Actual Comments:} \ding{182} So did the children's English grades improve? \ding{183} Haha! It's truly a ``combination''  of virtue and art! \ding{184} Teaching by example?\newline{}\textbf{Generated Comments:} \ding{182} This is too scary, how could there be such a teacher? This is not education. This is a crime! I hope such behavior can be severely punished by the law and we can protect children's safety! \ding{183} This is completely a moral decay, and the law must not be lenient. \ding{184} This is so shocking! As a teacher, it is truly irresponsible for students and society to use one's position to do such a thing. \\
    \bottomrule
    \end{tabular}
\end{table*}%

\section{Case Analysis}
We analyze three cases from the testing set, aiming to know the individual usefulness of the generated comments rather than the compound effects of generated comments and feedback understanding (Table~\ref{tab:cases}). The predictions are from dEFEND, with actual or mixed (actual and generated) comments provided.
In Case 1 (real news), actual comments are mostly of a questioning tone and thus not diverse, probably causing the model's misjudgment.
In contrast, the generated comments provided additional positive feedback, helping correct the prediction. 
This indicates that a partial observation of users' reactions might negatively impact distinguishing the fake from the real, especially when comment distribution is overly biased.
Differently, the fake case (Case 2) sparks more generated comments of doubtful and questioning tones, helping the model finally correct the prediction. The two cases confirm the potential of LLM-generated comments in helping detection, especially when actual comments are not diverse or even unavailable.
Case 3 is a failure case. We attribute the failure to the limited coverage of user responses, mostly about expressing anger, sarcasm, and disgust. We did not find generated comments that question the underlying news intent, which is recently considered important for this task~\cite{intent}.

\section{Conclusion and Discussion}
We proposed using large language models as news comment generators for improving comment-based fake news detection, especially when only a limited number or even none of the actual comments are available.
To obtain diverse comments and analyze them more comprehensively, we designed the \textbf{GenFEND} framework to enhance fake news detection with generated comments, which features multi-subpopulation feedback understanding and intra- and inter-view aggregations.
Experimental results on three public datasets demonstrated GenFEND's effective enhancements for content-only and comment-based fake news detection.
Further analysis reveals that GenFEND's effectiveness was derived from the broader coverage of potentially active and silent users and the multi-view multi-subpopulation analysis on generated comments.

\textbf{Discussion.} Technically, GenFEND exemplified how synthetic data empowers fake news detection. 
We identified the unnoticed and probably unavoidable limitations of existing detectors using actual user comments. As a substitute, we carefully designed a solution that elicits LLMs' capabilities in understanding natural language and playing specific roles for useful data synthesis. Despite this, we \textit{do not} intend to do extremely real simulations of actual cases but to provide an \textit{educated imagination} of discussions among different types of users. This idea allows us to leverage the advantages of LLMs and meanwhile avoid their disabilities (e.g., hard to speak as informally as social media users after alignment). Though our solution is domain-specific, it can serve as a useful reference for developing data-centric solutions for other social media tasks.
From the application view, this research again exhibited the dual role of LLMs~\cite{lucas-etal-2023-fighting}: They are not only misinformation creators but can also serve as defenders against misinformation.

\textbf{Limitation and Future Work.} 
We identify the following limitations of this paper:
\textbf{(1) }Besides gender, age, and education views, other demographic factors might also be useful for diversifying generated comments. We plan to extend and test more factors in the GenFEND development.
\textbf{(2) } Though generated comments have been shown helpful in fake news detection, they are not always more effective than actual comments (e.g., dEFEND prefers actual ones), indicating that rooms still exist in finding better strategies to utilize generated comments.
\textbf{(3) }Due to privacy concerns, we identify potentially active user types by inferring the profiles of actual commenting users, which may bring errors.
\textbf{(4)} We conducted experiments on two well-recognized yet costly API-based LLMs to ensure comment comprehensiveness based on their good instruction-following capability but did not test open-source models like LLaMA~\cite{llama2}, ChatGLM-3, and Gemma~\cite{gemma}. We plan to test more deployable LLMs at a lower cost.


\balance
\bibliographystyle{ACM-Reference-Format}
\bibliography{sample-base}

\end{document}